# Nondestructive Quality Control in Powder Metallurgy using Hyperspectral Imaging

Yijun Yan, *Member, IEEE*, Jinchang Ren, *Senior Member, IEEE*, He Sun

*Abstract*: Measuring the purity in the metal powder is critical for preserving the quality of additive manufacturing products. Contamination is one of the most headache problems which can be caused by multiple reasons and lead to the as-built components cracking and malfunctions. Existing methods for metallurgical condition assessment are mostly time-consuming and mainly focus on the physical integrity of structure rather than material composition. Through capturing spectral data from a wide frequency range along with the spatial information, hyperspectral imaging (HSI) can detect minor differences in terms of temperature, moisture and chemical composition. Therefore, HSI can provide a unique way to tackle this challenge. In this paper, with the use of a near-infrared HSI camera, applications of HSI for the non-destructive inspection of metal powders are introduced. Technical assumptions and solutions on three step-by-step case studies are presented in detail, including powder characterization, contamination detection, and band selection analysis. Experimental results have fully demonstrated the great potential of HSI and related AI techniques for NDT of powder metallurgy, especially the potential to satisfy the industrial manufacturing environment.

## I. Introduction

Hyperspectral Imaging (HSI) is a technique that combines spectroscopy with digital imaging. Regular multispectral systems, such as RGB cameras, collect information in a limited number of distinct wavebands spread out over a certain spectral range. HSI in contrast captures intensities over a continuous spectral range in very narrow wavebands. Each pixel does not only represent spatial information in form of *x* and *y* coordinates but also spectral information in form of a continuous spectrum. Depending on the system, this can entail several hundred wavebands. The data is stored in a three-dimensional data cube, often referred to as a hypercube where for each wavelength, a full resolution spatial image is available.

Traditional applications of HSI are in the field of remote sensing such as land mapping [1], food quality monitoring [2] and medical applications [3], etc. Due to recent advances in imaging technology in the last decades, HSI became more popular for industrial applications such as peatness prediction in the malted barley for Whisky manufacturing [4], non-destructive testing for End-of-Life carbon fiber reinforced polymer remanufacturing [5]. The popularity derives from the non-destructive nature of HSI, where samples can be analyzed chemometrically without altering their physical integrity. A second advantage is the rapid data acquisition. HSI data can be acquired in real-time and the subsequent data analysis is subject to the efficiency of algorithms and the computational power of the host system. As a result, HSI poses the great potential of a real-time chemometric analysis tool that can seamlessly be integrated into the processing chain of industrial production.

Taking advantage of 3D printing, additive manufacturing has become one of the most significant manufacturing industries in the United Kingdom due to its capacity to save time and money, decrease waste, reuse waste material during the printing process, and streamline supply chains. Now, collaboration between additive manufacturing (AM) and other major industries such as aerospace, medicine, automotive, architecture, etc., is progressively growing. In the next five years, the additive manufacturing market is projected to expand between $51 billion to $120 billion [6].

During the powder metallurgy process, a three-dimensional object will be formed through layering pure material. Typically, 5-10% of raw material in the powder bed will be used to make a component, which means the remaining raw material will be returned to the collector chamber for reuse in the manufacturing of another component. However, the reuse of metal powders brings contamination risks, where numerous processes are required to ensure the purity of the material. In addition, due to long time reuse procedure, some degradation such as creep, corrosion, oxidation, etc., might be introduced into the material, which leads to the change of its composition and size distribution. Therefore, condition assessment in additive manufacturing is crucial because impurities in the raw material can lead to cracking and malfunctions. Material contamination might be deadly in applications such as aircraft and medical devices. To this end, a reliable and efficient non-destructive condition assessment technique is highly needed to inspect the nature of metal powder materials and the manufacturing process for improved quality control [7].

Current popular inspection methods for powder metallurgy include visual inspection, ultrasonic testing, thermal imaging, X-ray computed tomography (XCT) and metallography[8, 9].

Visual inspection is a quick and low-cost inspection method which doesn't rely on extra instrumentation. However, it is less effective for metallurgical condition assessment due to two main reasons. First, the popular metal powders including Titanium, Aluminium, Tungsten, Steel, and Nickel have similar colour properties. Second, their particle size is micron-level. As

We greatly appreciate the funding support from the Carpenter Additive, LPW Technology Ltd, who also provided the metal samples in every case studies.
Corresponding Author: Prof. J. Ren (jinchang.ren@ieee.org).

Y. Yan and J. Ren are with National Subsea Centre, Robert Gordon University, Aberdeen, U.K.
H. Sun is with School of Computer Science and Technology, Beijing Institute of Technology, China.

a result, the contamination cannot be effectively detected by the naked eye and inspection techniques that depend on colour.

Ultrasonic testing has been widely used to inspect internal defects of metal components. It employs a transmitting probe that transmits ultrasonic waves through the component and then emit various reflected signals when encountering different interfaces. Transmission time difference can reveal faults like pores and cracks [10]. Recently, laser ultrasonic inspection has been developed for real-time AM monitoring. It generates the ultrasonic waves by laser pulse and detect the reflected signal by laser interferometer. However, ultrasonic testing struggle with non-smooth and complicated surfaces. Thus, most of the workpieces surface needs to be treated to eliminate the influence of surface roughness [11].

Thermal imaging is mostly carried out by infrared cameras for temperature measurement. Any physical changes and defects during the manufacturing will affect the heat conduction in the workpieces, and present a higher temperature response in infrared images [12]. However, making accurate measurement relies on high-sensitive and expensive infrared camera [13]. A comprehensive calibration must be performed by experienced technician depending on the ambient conditions and materials.

XCT is an imaging technique that reconstructs the 3D structure of an object by capturing several X-ray pictures around a rotating axis. It can expose the internal structure of a metallurgical product, allowing for the detection of flaws such as cracks, porosity, inclusions and density variations [14]. Since XCT relies on x-ray penetration of the object to be scanned, larger size of the object will reduce the maximum possible magnification of the scan which leads to a decreased spatial resolution [15]. Although this issue can be mitigated by stitching multiple locally tomographic images leading to much larger tomograms, the results can be still skewed and make detecting small defects a challenge [16].

Metallography is also known as optical microscopy. Several 2D images will be captured by optical light microscope and then stitched together to create and record the whole cross section of the workpieces [17]. This method can provide very high precision for pore recognition. However, its experimental setup and calibration are very complex. For example, the selection of lens magnification and the number of captured images under different magnification level need to be carefully decided according to specific needs. Also, manual focus adjustment and location choice during the data capturing may also cause human errors and increase uneconomical manpower [18].

However, most of these techniques extract information about the physical integrity of the as built structures rather than the material composition from the metal powders. In this context, HSI can provide a novel insight to fill this gap. It will not be limited by environmental temperature and geometry of the object, making it a unique solution for non-destructive inspection far beyond conventional techniques. It is expected that this will subsequently reduce the costs and environmental impact associated with the production of unqualified products.

The aim of this study is to explore the feasibility of HSI for the contamination estimation in metal powders, thus narrowing the gap between academic technique and industrial application. To achieve this, three experiments including the characterizing of metal powders, detection of contamination, and band selection are carried out. Details of sample generation will be discussed in the Experiment Setup Section.

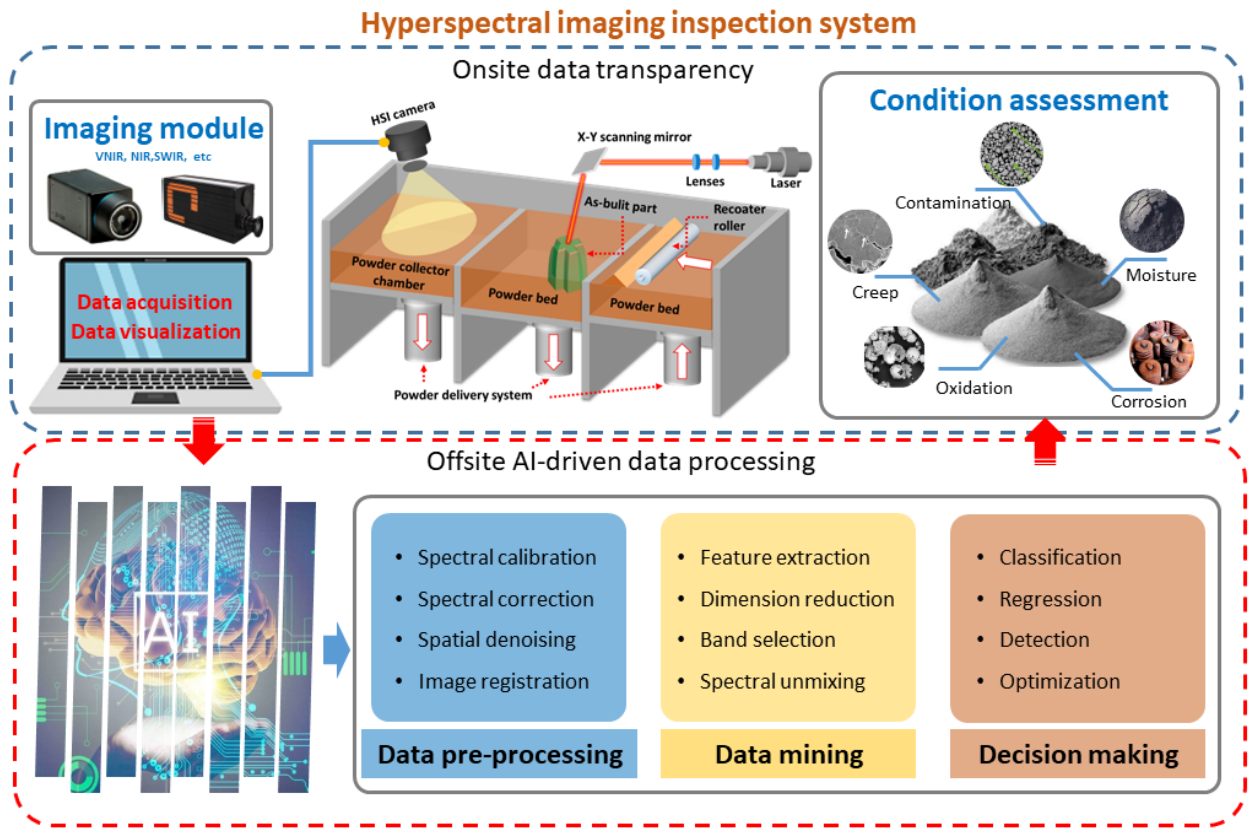

Fig. 1. Hyperspectral imaging inspection system for powder metallurgy.

The main contributions of this paper are highlighted below.
1. We propose an HSI-based fast, accurate and non-destructive inspection system for contamination detection of metal powder in the additive manufacturing.
2. An extendable framework of data processing is proposed for efficient measurement of powder characteristics.
3. We have carried out step-by-step case studies to validate the feasibility and effectiveness of the HSI based inspection system in the additive manufacturing, where useful discussions and conclusions were reported.

The remaining paper is organized as follows: Section II describes the concept of the HSI based inspection system. Section III discusses the way to data acquisition and data processing in this system. Section IV details the three case studies. Finally, some remarkable conclusion and future directions are summarized in Section V.

## II. HSI Inspection System

Fig. 1 demonstrates the concept of the HSI inspection system for additive manufacturing, which is composed of two interactive modules, i.e., Onsite data transparency and Offsite AI-driven data processing. Onsite data transparency includes imaging module and condition assessment. In the imaging module, the optical sensors such as VIS, NIR, and short-wave infrared (SWIR) hyperspectral cameras can provide a continuous spectral information over a certain spectral range, along with a good spatial resolution. Then the acquired information flow will be transferred from onsite to offsite for data processing and analysis.

Within the data processing techniques, three core steams are covered, i.e., Data pre-processing, Data mining and Decision making. Data pre-processing is a vital step after data acquisition for data calibration and enhancement, which can produce high-quality data for easy understanding and improved analysis via data visualization and analytics. In general, spectral calibration is an essential step to normalize the captured spectrum. According to the practical needs of different NDT tasks, spectral correction and spatial denoising sometimes needs to be applied for improving the data quality. For some large object, image registration and stitching are needed to tackle with the limitation of the field of view and cove the whole surface.

The main data mining techniques for HSI data include but not limited to spectral-spatial feature extraction, dimensional reduction, band selection, and spectral unmixing. One of the biggest issues with analyzing hyperspectral data is the large number of variables involved, leading to the curse of dimensionality or Hugh's phenomenon. A high number of variables necessitates a considerable number of samples thus the corresponding memory and processing capacity, and may potentially result in overfitting of the classifiers during model training. To address this issue, feature extraction is needed for dimension reduction and more effectively data characterization.

Machine learning algorithms such as Support Vector Machine (SVM), Random Forest (RF) and Neural Network (NN) have been widely applied in numerous industrial applications [4] [5]. The integration of data pre-processing, feature extraction and machine learning can provide an automatic vision-based inspection, which is usually built upon the sample data (also known as training data) and eventually used for decision making such as classification, regression, detection and optimization.

Finally, the outcome from AI-driven decision making will be useful for detecting the defects such as contamination, moisture, corrosion, oxidation and creep in the metal powder. In summary, the proposed hyperspectral imaging inspection system can be employed for a better condition assessment of metal powder, which will help the onsite operators make more efficient manufacturing strategies to improved quality control in powder metallurgy.

## III. Materials and Methodologies

In this section, the standard progress of data acquisition in the lab environment and data pre-processing techniques are introduced. Those techniques are not only used in this work but also applicable in many other HSI applications.

### A. Data acquisition

In our three case studies, two HSI imaging devices including a near-infrared (NIR) camera and a visible-near infrared (VIS) camera are used for data acquisition. Both VIS and NIR cameras operate in a push-broom mode, meaning that the camera is pointed downwards and scans only a single line at a time. Each scan produces a two-dimensional image in which one dimension represents the spatial line and the other dimension represents the complete spectrum of each pixel. The objects to be scanned are moved with a translational stage at even speed underneath the camera and are thereby fully scanned. This means that the only spatial limitation is the width of the objects. They can be theoretically infinitely long, only limited by the storage capacity. The camera settings in the lab environment are shown in Fig. 2. For every case study, two 20 W Tungsten halogen lights are used for illumination. The samples to be scanned are moved using a translational stage at a consistent speed of 16mm/s with a working distance of 25 cm beneath the camera, to form the 3D hypercube data.

The VIS imaging system is comprised of a Hamamatsu ORCA-03G CCD camera and a Specim V8E spectrograph. This system operates in the spectral range of 400-950 nm and has a spectral resolution of 2nm. In addition, 4-fold spatial and spectral binning were applied in order to minimize noise and boost the camera's light sensitivity. This also results in an image with 336 pixels per line and 256 spectral responses per pixel.

The NIR imaging system is Innospec Red Eye 1.7. It operates in the spectral range of 950-1700 nm with a spectral resolution of 10 nm. The number of spectral bands and spatial pixels per line without binning is 256 and 320, respectively.

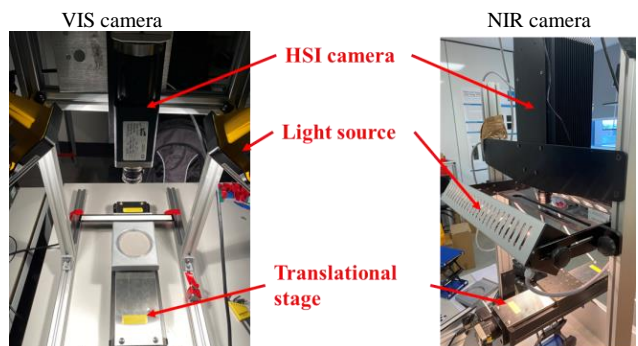

**Fig. 2. Camera settings in the lab environment**

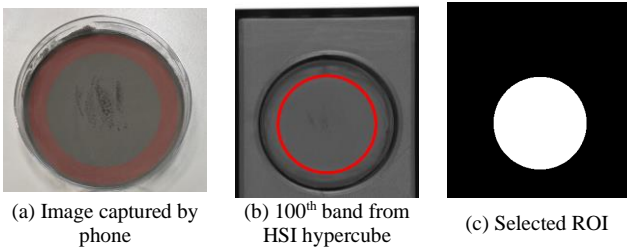

(a) Image captured by phone   (b) 100th band from HSI hypercube   (c) Selected ROI

Fig. 3. Illustration of ROR extraction.

For each sample, we spread out the metal powder in a round container. Five pure samples are used for training and the mixed samples are used for testing. To eliminate the effects of the ambient light and inconsistency between the training and testing dataset, each container of metal powder is scanned 4 times where each container is rotated in 4 different directions (0^°, 90^°, 180^°, 270^°) to form four hypercubes as training or testing datasets. The obtained 4 hypercubes will be pre-processed and stacked in the pixel-wise manner to produce a stacked image data pool for training or testing.

### B. Data processing

#### 1) Spectral calibration

During the data acquisition stage, the lighting conditions may shift within a hypercube or apparently between different datasets along the scan lines. To mitigate the effect of camera quantum and physical configuration differences on imaging systems, reliable calibrations for the HSI system are required to ensure the stability and acceptance of the hyperspectral data produced. As a result, light calibration is required to convert the raw radiance spectrum $s$ to the reflectance spectrum $r$ in order to reduce this incoherence and retain a consistent influence of the light conditions. Without exposing the camera to light, we may get a dark reference spectrum $d$. Then a white reference spectrum $w$ can be obtained by imaging an ideally reflective white surface (e.g., Spectralon with Lambertian scattering). The present illumination's light sensitivity may be calculated by normalizing the signal by:

$$r = \frac{s - d}{w - d} \quad (1)$$

#### 2) ROI extraction

During the sample preparation, the metal powders are spread out in the round container. After data acquisition, the HSI data will have the spectra of metal powder and the background. As our research focus is the characterization of metal powder, in this section, we employ a robust circle detection method to extract the pixels of metal powder only and remove the shadow that usually exists near the container border (highlighted by red ring in Fig. 3(a). This will speed up the following on data mining and improve the accuracy of classification. Then, every sample will conduct a data matrix with N=22797 spectral samples.

The process of circle detection method is summarized in the following steps:
1. Apply an adaptive thresholding method [19] on the 100th band image to extract a binary template;
2. Calculate its centroid using central moment, i.e. the mean x and y value of all white pixels;
3. Determine the minimum radius to generate the circular mask, as seen in Fig. 3(b);

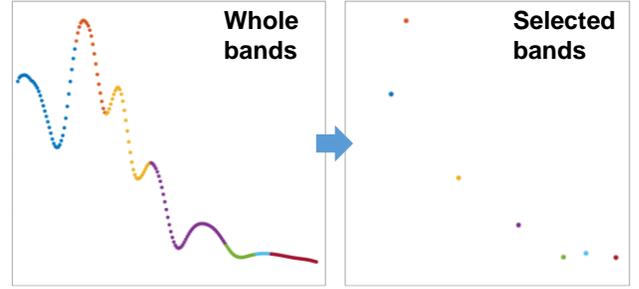

Fig. 4. Illustration of band selection

4. Apply the mask on the pre-processed hypercube to extract the pixels in red circle as our extracted ROI (Fig. 3(c)).

#### 3) Band selection

Thanks to the advancement of imaging spectrometer technology, most off-the-shelf hyperspectral sensors can provide hundreds of spectral bands and each of them records the reflectance of objects to a specific electromagnetic wave. Although hyperspectral data provides rich information, high dimensionality is still a major trouble of HSI. The resulting high computational cost makes fast onsite processing and practical deployment difficult[20].

To reduce the dimension of HSI data but keep its representative information, feature extraction and feature selection (also known as band selection) techniques have been widely explored for many years. Feature extraction techniques usually project the entire HSI data into a new low-dimensional feature domain to extract the discriminative information. However, it will bring the difficulty to data interpretation as the physical meaning of entire HSI data has been destroyed. In this case, band selection can be more useful as it only drops the redundant bands and preserve the dominant bands (Fig. 4).

### IV. CASE STUDIES AND DISCUSSIONS

In this section, three case studies are presented to show the potential of proposed HSI based NDT of metal powders, including specific AI techniques introduced in the context.

### A. Case study 1: Conceptual validation

In this study, we assume the attributes of pure samples spread out in each container are consistent. Therefore, HSI can well characterize the pure metal powders.

#### 1) Experimental materials and settings

To validate our hypothesis, five categories of pure metal powders including Titanium (TI64GD23-ACFP), Aluminium (AISi10Mg-F), Tungsten (LPW-W-AAAC), Steel (316L-F), and Nickel (718-F), are used to generate five pure samples in this study. Here, we consider this study as a classification problem. Therefore, a widely used classification model, Support Vector Machine (SVM), is employed for data classification, and the performance is quantitatively evaluated using the Overall Accuracy (OA).

$$OA = \frac{1}{N} * \sum_{i}^{T} C_i * 100\% \quad (2)$$

where $N$ is the number of observations, $T$ is the number of classes, $C_i$ represents the number of correctly classified observations in class $i$.

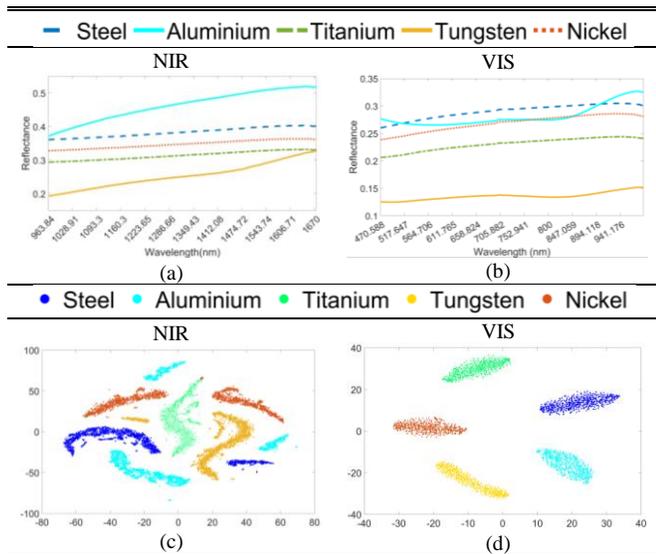

**Fig. 5. Spectral profile (a, b) and t-SNE feature space (c, d) of five pure samples based on NIR and VIS imagery.**

There are three main reasons to choose SVM as the classifier. First, it can exploit a margin-based criterion and is very robust to the Hughes phenomenon [21]. Second, it can produce better results than other classifiers and has been widely used for HSI classification [1, 4]. Third, there are several available libraries supporting multiple functions of SVM, where we use the LIBSVM [22] for implementing multi-class classification.

Each experiment was repeated 10 times and the average results are reported to provide the statistical significance and avoid systematic errors. Within each repetition, training and testing samples are disjoint. Different numbers of pixels (i.e., 5, 10, 15, 20, 25, 30, 5%, 10%) in each category have been randomly selected for training.

*2) Results and discussion*

To visualize the spectral characteristics of five pure samples, the average spectral profiles over the whole bands of NIR and VIS are shown in Fig. 5 (a & b). Aluminum and Tungsten have quite distinct spectral profiles, but Steel, Titanium, and Nickel have a fairly similar trend, despite their varying intensities. To further visualize the discrimination ability of NIR and VIS imagery, t-distributed stochastic neighbor embedding (t-SNE) method [23] was applied to map the whole spectral band to the 2D feature space. As seen in Fig. 5 (c), although two or more clusters of individual samples are shown at different locations, the t-SNE feature space correctly separates the boundaries between each sample acquired by NIR imagery. The extracted features of each sample obtained by VIS imagery tend to be organized into distinct clusters, indicating the great discriminability of the VIS spectrum.

Table 1 Contents in each mixed sample

| ID | Description | |
|----|-------------|---|
| 1 | Titanium (≥99%) + Tungsten (≤1%) | Shaken |
| 2 | Nickle (≥99%) + Aluminium (≤1%) | Shaken |
| 3 | Steel (≥99%) + the rest four pure samples (≤1%) | Shaken |
| 4 | Aluminium (≅70%) + Nickel (≅30%) | Shaken |
| 5 | Tungsten (≅70%) + Titanium (≅30%) | Shaken |
| 6 | Steel (≅70%) + Aluminium (≅30%) | Shaken |
| 7 | Titanium (≥99%) + Tungsten (≤1%) | unshaken |
| 8 | Titanium (≅20%) + Aluminium (≅20%) + Tungsten (≅20%) + Steel (≅20%) + Nickel (≅20%). | unshaken |

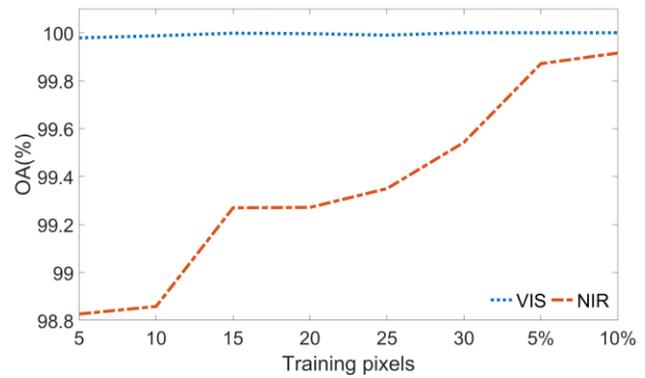

**Fig. 6. Classification results with multiple training rates.**

The classification results of NIR and VIS imagery under 8 training rates are shown in Fig. 6. As seen, the classification accuracy of VIS and NIR grows with the increasing of the training rate. Due to higher discriminability, VIS imagery with an OA approaching 100% appears to perform better than NIR imagery. Nevertheless, the OA of NIR imagery can still maintain in a very high level (>98.8%). In this context, HSI has been validated for characterizing the pure metal powders.

*B. Case study 2: Contamination detection*

In this study, we assume HSI is capable of detecting the contaminations in the mixture metal powders.

*1) Experimental materials and settings*

To simulate the contamination scenario, eight mixed samples were created from five raw samples. The details of the mixed sample are reported in Table 2, where shaken and unshaken indicate whether or not the container was manually shaken after metal granules were added. After that, we carefully place the containers to be scanned on the movable system to prevent secondary powder mixing caused from undesired shaking.

In this study, the identification of contamination can be considered as an anomaly detection problem. To discover the anomalies in the mixed samples, a unary classification alike technique is utilized instead of the standard cross validation training strategy. In specific, all pixels of five pure samples are used as training pixels to train the SVM model which will then be used to detect and recognize the anomalies in the mixed samples to be tested.

*2) Results and discussion*

The anticipated proportion of each type of metal powder in the combined samples is displayed in Table 3. Based on the findings of the 1st and 2nd mixed samples, it appears that VIS spectral bands are more useful than NIR spectral bands for identifying contamination in a very small percentage. For the 3rd mixed sample, NIR imagery can estimate a reasonable portion of dominant metal powder i.e., Steel, and recognize the

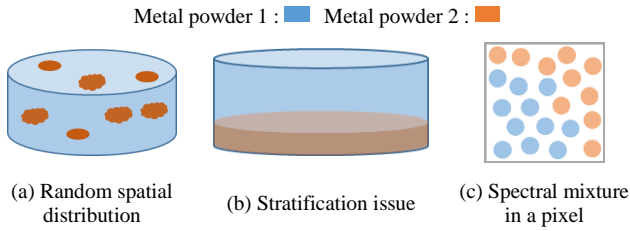

(a) Random spatial distribution  (b) Stratification issue  (c) Spectral mixture in a pixel

**Fig. 7.** Illustration of the reasons for misclassification and failure issue

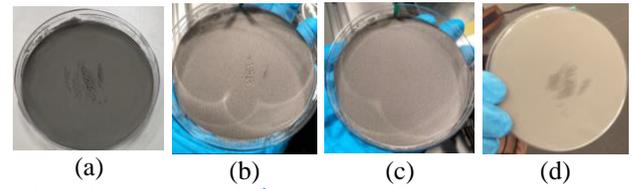

(a)   (b)   (c)   (d)

**Fig. 8.** Surface condition of 7th sample, (a) after sample preparation, (b) after several minutes, (c) in the final state. (d) Bottom condition of 7th sample in the final state.

Aluminium and Nickle as contamination but fail to detect the Titanium and Tungsten. VIS imagery can also detect the Steel, Aluminium and Nickle in this mixed sample, however, the estimation of Aluminium and Nickel is far from anticipation.

Possible explanations for misclassification and failure in these three mixed samples are:
1. After shaking the container, two classes of powder will be mixed up. Then the classification performance mainly depends on the way to shaking, which is a random scenario (Fig. 7(a)).
2. In general, large density particles tend to settle in the bottom layer ((Fig. 7(b)), making identification challenging.
3. Single pixels contain two or more classes of powder particles, and the classifier has low discriminative ability of mixed spectrum ((Fig. 7(c)).

To validate these three possible reasons, five more mixed samples are generated. For the 4th – 6th mixed samples, the proportion of contamination was raised to 30%. For the 4th mixed sample, NIR imagery can offer an accurate estimation of Aluminium, but incorrectly identifies Nickel as Steel. This is because the main physical properties of Steel and Nickle in this work are comparable, resulting similar trend of spectral profile, particularly in the NIR band (Fig. 5(a)). VIS imagery can recognize the Nickle as the main contamination, although the estimated proportion is significantly below expectations. For the 5th mixed sample, both VIS and HSI imagery can recognize the predefined contamination, but VIS can provide a more accurate assessment. For the 6th mixed sample, both NIR and VIS imagery can identify the Steel and Aluminium but fail to estimate the expected portion of predefined metal powders. The most likely explanation behind that is because the unpredictable spatial distribution of metal powders impedes precise estimation. On the other hand, the problem of spectrum mixing, which is also caused by the spatial resolution restriction of HSI cameras, is seen as the primary cause of misclassification.

To mitigate the unpredictable spatial distribution issue, 7th and 8th samples are generated. For 7th sample, we first spread the Titanium over the bottom of the container, then carefully spread tiny portion of Tungsten on top of Titanium. For 8th sample, five pure metal powders are spread over the container in turn. Finally, we place both containers for scanning without shaking. From the results of 7th sample, several findings can be concluded. First, both VIS and NIR imagery can identify Titanium as main element and Tungsten as a contamination. Second, misclassification still happens to VIS imagery, which is again caused by spectral mixture issue. Although the container hasn't been shaken, some Titanium and Tungsten particles on the surface are still mixed in the pixel level, which mislead the classifier to make the wrong estimation. On the other hand, NIR imagery yields far more plausible estimation. The reason is because the 7th sample is first scanned by VIS cameras and then by NIR camera, and the surface condition of metal powders changes during the transition from one camera to another. As particles with greater density tend to settle at the bottom, Tungsten particles on the surface will be underneath of the Titanium particles. As seen in Fig. 8 (a), we can clearly see the Tungsten powders are in the middle of Titanium powders when the mixed sample was freshly created. Several minutes later, the majority of the Tungsten particles have passed through the Titanium powders, while a few remain on the surface (Fig. 8 (b)). Eventually, only Titanium particles will be visible on the surface (Fig. 8 (c)), while Tungsten granules will completely settle at the bottom of the container (Fig. 8 (d)). This situation may also occur with other samples such as 4th sample where the density of Nickle is larger than that of Aluminium, which explains misclassification from a different perspective. In this case, spectral mixture issue is no more existing, but such condition still brings the difficulty of contamination detection.

From the results of the 8th sample, both NIR and VIS imagery can well identify the main elements in the same container despite the fact that the estimated portion of each metal powder doesn't match the ideal value because of the overlapping issue.

In summary, all these results actually validate that HSI is able to recognize the contamination in the mixed sample under two preconditions, i.e., metal powders to be characterized need to be on the surface and minimum portion of metal particle needs be larger than a pixel.

**Table 2** Results of case study 2. The dominant metal powder and contamination are highlight in bold and shading italic, respectively

| ID | | Steel | Aluminium | Titanium | Tungsten | Nickel | | Steel | Aluminium | Titanium | Tungsten | Nickel |
|---|---|---|---|---|---|---|---|---|---|---|---|---|
| 1 | NIR imagery | 0.000 | 0.000 | **100.000** | *0.000* | 0.000 | VIS imagery | 0.000 | 0.000 | **99.959** | *0.041* | 0.000 |
| 2 | | 0.225 | *0.000* | 0.961 | 0.000 | **98.824** | | 3.108 | *0.004* | 0.179 | 0.000 | **96.709** |
| 3 | | **99.815** | *0.018* | 0.000 | 0.000 | *0.167* | | **69.703** | 7.437 | 0.000 | 0.000 | *22.860* |
| 4 | | 27.950 | **70.646** | 0.000 | 0.000 | *1.405* | | 0.321 | **96.355** | 0.033 | 0.000 | *3.291* |
| 5 | | 0.000 | 0.000 | *4.600* | **95.400** | 0.000 | | 0.000 | 0.000 | *28.181* | **71.819** | 0.000 |
| 6 | | 25.760 | 26.020 | 8.250 | 0.000 | 36.160 | | 25.020 | 65.740 | 0.21 | 0 | 3.12 |
| 7 | | 0.000 | 0.000 | **99.627** | *0.328* | 0.046 | | 0.002 | 4.434 | **86.518** | *1.191* | 7.857 |
| 8 | | 13.347 | 18.215 | 25.036 | 13.169 | 30.233 | | 19.170 | 20.483 | 19.700 | 12.558 | 28.088 |

Table 3 Estimation results of ADBH on the 7th sample, where exp. represents expectation.

| No. Band | Element | Steel | Aluminium | Titanium | Tungsten | Nickel | MAE | Element | Steel | Aluminium | Titanium | Tungsten | Nickel | MAE |
|---|---|---|---|---|---|---|---|---|---|---|---|---|---|---|
| | Exp. | 0.00 | 0.00 | 99.00 | 1.00 | 0.00 | | Exp. | 0.00 | 0.00 | 99.00 | 1.00 | 0.00 | |
| | Baseline | 0.00 | 0.00 | 99.63 | *0.33* | 0.05 | | Baseline | 0.00 | 4.43 | 86.52 | *1.19* | 7.86 | |
| 3 | NIR imagery | 0.00 | 0.00 | 99.63 | 0.28 | 0.10 | 0.021 | VIS imagery | 0.00 | 0.66 | 92.75 | 1.27 | 5.32 | 2.525 |
| 4 | | <span style="color:red">0.00</span> | <span style="color:red">0.00</span> | <span style="color:red">99.62</span> | <span style="color:red">0.29</span> | <span style="color:red">0.09</span> | <span style="color:red">0.018</span> | | 0.00 | 1.68 | 93.73 | 1.34 | 3.24 | 2.947 |
| 5 | | 0.00 | 0.00 | 99.60 | 0.28 | 0.11 | 0.028 | | 0.00 | 1.95 | 94.15 | 1.25 | 2.65 | 3.077 |
| 6 | | 0.00 | 0.00 | 99.62 | 0.29 | 0.09 | 0.018 | | 0.00 | 2.29 | 92.61 | 1.23 | 3.87 | 2.453 |
| 7 | | 0.00 | 0.00 | 99.60 | 0.30 | 0.11 | 0.024 | | 0.00 | 3.23 | 90.67 | 1.19 | 4.91 | 1.661 |
| 8 | | 0.00 | 0.00 | 99.62 | 0.28 | 0.10 | 0.022 | | 0.00 | 3.00 | 91.10 | 1.23 | 4.66 | 1.851 |
| 9 | | 0.00 | 0.00 | 99.61 | 0.29 | 0.10 | 0.022 | | 0.00 | 5.75 | 86.97 | 1.21 | 6.07 | 0.715 |
| 10 | | 0.00 | 0.00 | 99.61 | 0.29 | 0.10 | 0.022 | | <span style="color:red">0.00</span> | <span style="color:red">4.46</span> | <span style="color:red">88.24</span> | <span style="color:red">1.15</span> | <span style="color:red">6.14</span> | <span style="color:red">0.702</span> |

### C. Case study 3: Band selection

In this study, we assume hyperspectral band selection is able to reduce the redundancy in the hyperspectral data whilst maintaining the contamination detection performance.

*1) Experimental materials and settings*

To assess the efficacy of band selection techniques, we select three typical samples (i.e., 3rd, 5th and 7th) from case study 2, and choose four cutting-edge band selection methods i.e., ADBH [24], OCF [25], ONR [26], KNGBS [27]. Then, taking the experimental results from the second experiment as a baseline, we compare the detection performance before and after adding these band selection methods in terms of mean absolute error (MAE). Lower MAE value indicates that the selected bands can make closer estimation of contamination to what entire bands did.

*2) Results and discussion*

An example of investigating ADBH method on the 7th sample is shown in Table 4, where the estimated results that are most approaching to the baseline is highlighted in red. For NIR and VIS data, selecting four and ten spectral bands can give us the most comparable estimation to the baseline, respectively. However, it is worth noticing that the estimated results with 5 selected bands in VIS imagery seems to be much closer to the practical expectation. The possible reason is the entire spectral bands may contain more redundancy than useful information, resulting in biased estimation. To avoid monotonous and keep the discussion tidy, MAE value of four band selection methods on three samples is calculated and presented in Fig. 9. For the 3rd sample, OCF can produce the promising result when extracting 10 bands from VIS data, while ADBH performs better on NIR data with only 5 selected bands. For the 5th sample, the lowest MAE happens when 6 bands are selected by ADBH for VIS data and 7 bands are selected by ONR for NIR data. For the 7th sample, OCF is considered to be the best method for VIS data as it can select the most representative three spectral bands and get the promising results. Meanwhile, FNGBS works the best for NIR data by selecting 9 bands. On a different note, higher volume of bands doesn't benefit the MAE evaluation necessarily. For example, MAE of ABDH varies with the increasing number of bands in NIR data of 3rd sample, MAE of OCF rises with the growth of band number though it drops down when the number of bands equals to 9.

All of these results and findings have demonstrated the immense potential of band selection on HSI data to facilitate more effective and efficient decision-making. The practical performance may vary according to the sample conditions and adopted band selection methods. Although exhaustive studies are required to attain the optimal band volume and optimum decision-making outcomes, only a few spectral bands can still provide satisfactory outcomes in most cases.

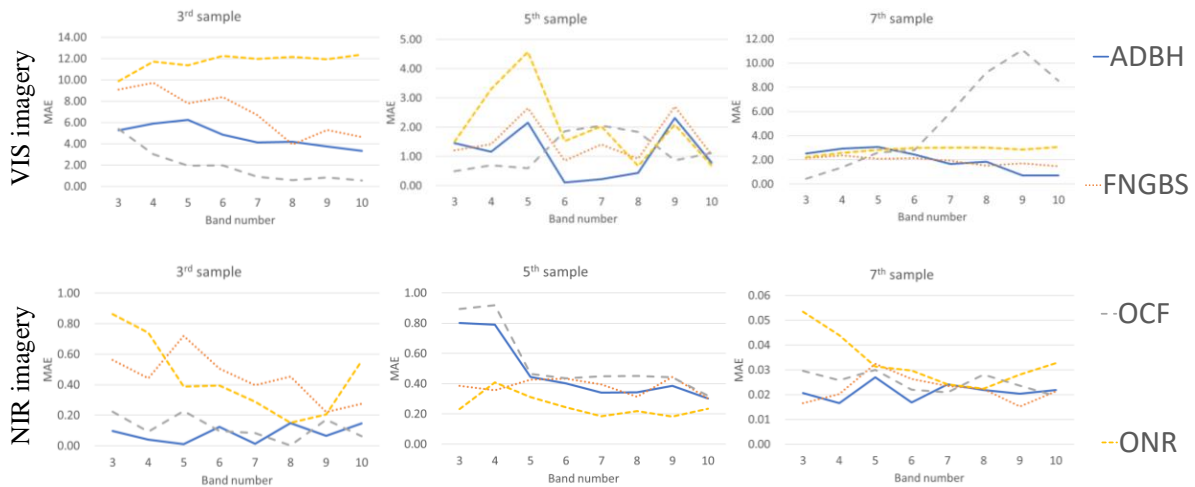

**Fig. 9.** Mean absolute error (MAE) of band selection methods on three samples.

## V. Conclusions and future work

In this paper, three case studies are used to offer unique insights on HSI-based NDT of powder metallurgy. Five categories of pure meter powders were employed to generate 13 mixed samples in total for validating the pre-assumptions. The experimental results have demonstrated the great value of the integration of machine learning and HSI for quality control of metal powder in real applications of additive manufacturing, with particularly promising results in the identification of contamination and material characterization. Some remarkable findings are drawn below:

1. Our proposed NDT framework can detect the contamination and characterize different mixtures of metal powders even at a lower ratio of 1% under two preconditions: 1) the particle group is expected to be large enough for a pixel to capture, otherwise the spectra mixture will heavily affect the classification accuracy; 2) all classes of powder are expected to be on the surface. Any class of powder layer beneath another will result in misclassification.
2. Band selection is found to be useful to reduce the data redundancy and also produce satisfied estimation results with potential multispectral system. However, the estimation results may vary depending on the sample conditions and the adopted band selection methods.

In our future work, three main tasks will be conducted to mitigate the current constraints and further improve the reliability and feasibility of proposed system. First, to overcome the first precondition, hyperspectral unmixing methods can be employed to solve the spectra mixture problem when the spatial resolution is limited. Second, the restriction of HSI is that it cannot completely access the subsurface condition, which is also the primary reason for the second precondition. This constraint can be overcome by integrating HSI with other NDT techniques such as thermography. Third, a more comprehensive evaluation of state-of-the-art band selection methods will be examined. Guided by those results, a bespoke multispectral system for powder metallurgy can be developed for practical deployment. Apart from these studies, we will also explore new NDT tasks for improved adaptiveness and completeness to meet the ever-changing market requirements.

Last but not the least, there are two challenges in this new NDT technique, i.e., ineffective data annotation and inefficient data capturing. To solve the first challenge, a more detailed data quantification will be carried out, where an advanced spectrometer with higher spatial resolution can be adopted to capture higher quality data and assist the precisely data annotation. To tackle with the second challenge, snapshot hyperspectral cameras can be replaced with push-bloom ones to achieve the data acquisition in a timely manner.